\documentclass[runningheads]{llncs}
\usepackage[T1]{fontenc}
\usepackage{graphicx,verbatim}
\usepackage{amsmath}
\usepackage{amsfonts}
\usepackage{amssymb}
\usepackage{booktabs}

\usepackage{hyperref}

\usepackage{color}

\begin{document}
\title{Adaptation of Multi-modal Representation
Models for Multi-task Surgical Computer
Vision}
\titlerunning{MML-SurgAdapt}
%

\renewcommand{\thefootnote}{\fnsymbol{footnote}}
\footnotetext[1]{\textit{This manuscript has been accepted for publication and will appear in the proceedings of MICCAI 2025.}}

\author{Soham Walimbe \inst{1}\and Britty Baby\inst{1,2} \and
Vinkle Srivastav\inst{1,2} \and Nicolas Padoy \inst{1,2}}


%
\institute{University of Strasbourg, CNRS, INSERM, ICube, UMR7357, Strasbourg, France 
\and Institute of Image-Guided Surgery, IHU Strasbourg, Strasbourg, France \\ 
}
\authorrunning{S. Walimbe et al.}

\maketitle              
\begin{abstract}
Surgical AI often involves multiple tasks within a single procedure, like phase recognition or assessing the Critical View of Safety in laparoscopic cholecystectomy. Traditional models, built for one task at a time, lack flexibility, requiring a separate model for each. To address this, we introduce MML-SurgAdapt, a unified multi-task framework with Vision-Language Models (VLMs), specifically CLIP, to handle diverse surgical tasks through natural language supervision. A key challenge in multi-task learning is the presence of partial annotations when integrating different tasks. To overcome this, we employ Single Positive Multi-Label (SPML) learning, which traditionally reduces annotation burden by training models with only one positive label per instance. Our framework extends this approach to integrate data from multiple surgical tasks within a single procedure, enabling effective learning despite incomplete or noisy annotations.  We demonstrate the effectiveness of our model on a combined dataset consisting of Cholec80, Endoscapes2023, and CholecT50, utilizing custom prompts. Extensive evaluation shows that MML-SurgAdapt performs comparably to task-specific benchmarks, with the added advantage of handling noisy annotations. It also outperforms the existing SPML frameworks for the task. By reducing the required labels by 23\%, our approach proposes a more scalable and efficient labeling process, significantly easing the annotation burden on clinicians. To our knowledge, this is the first application of SPML to integrate data from multiple surgical tasks, presenting a novel and generalizable solution for multi-task learning in surgical computer vision. Implementation is available at: \url{https://github.com/CAMMA-public/MML-SurgAdapt}.

\end{abstract}
\section{Introduction}

AI models capable of handling diverse downstream tasks have gained increasing attention in computer vision research. Traditional task-specific models, designed for single tasks like multi-class classification, lack versatility and require separate architectures for each task, making them non-scalable and resource-intensive.  This limitation has spurred interest in scalable, task-agnostic models capable of generalizing across multiple tasks, including unseen classes \cite{yuan2024hecvl}.  Vision-Language Models (VLMs), like CLIP \cite{radford2021learning}, uses natural language supervision to align visual and textual representations in a shared semantic space. Trained on vast image-text datasets, VLMs excel in general tasks—such as object or scene recognition—via textual prompts and can be adapted through techniques like prompt learning \cite{zhou2022learning} or network adaptation \cite{yang2024mma}.

While VLMs thrive in everyday vision tasks, their application to surgical data science has been limited. Surgical AI involves analyzing complex video data from procedures, with tasks ranging from phase recognition (e.g., preparation) to domain-specific challenges like Critical View of Safety (CVS) assessment or action triplet identification (e.g., ‘forceps-grasp-gallbladder’). Traditionally, datasets such as Cholec80 \cite{twinanda2016endonet}, Endoscapes2023 \cite{murali2023endoscapes}, and CholecT50 \cite{nwoye2022rendezvous} have been curated for specific tasks, leading to specialized models \cite{ramesh2021multi,murali2023encoding,nwoye2022rendezvous} that address phase recognition, CVS prediction, and action triplet recognition independently. This task-specific approach, while effective in isolation, results in fragmented solutions that struggle to scale or integrate the complementary tasks inherent to a surgical procedure. To address this, we introduce MML-SurgAdapt, a framework that adapts a pretrained VLM (CLIP) using natural language supervision—pairing surgical images with discriminative text across multiple tasks—to create a unified, task-agnostic model. By merging Cholec80, Endoscapes2023, and CholecT50, we enable a cohesive approach that tackles these interconnected tasks together, responding to the need for scalable, integrated solutions in surgical AI that reflect the procedure’s holistic nature.

A key difficulty in this kind of multi-task learning is handling partial annotations, where each dataset labels only its target task, leaving others unlabeled. This creates a multi-label problem with potential false negatives. To address this, we turn to Single Positive Multi-Label learning (SPML), which was traditionally designed to streamline annotation in multi-label classification settings \cite{zhang2021simple}. In SPML, each image is assigned just one positive label, while all other potential labels are initially assumed to be negative, which can introduce false negatives. We mitigate false negatives using Hill loss \cite{zhang2021simple}, which re-weights negatives and outperforms other SPML losses in our experiments.



Our SPML-based approach enables multi-task integration even with incomplete data and enhance resilience against noisy or inconsistent labels. Interestingly, our implementation of SPML reduced the number of required labels by 23\%. This reduction is particularly significant for surgical tasks, where annotation complexity varies widely across labels. For instance, tasks like CVS assessment typically depend on time-intensive clinical reviews, which are costly and resource-heavy. By streamlining annotation demands, our framework enhances the practicality of multi-task learning for real-world surgical applications.

Our contributions include: (1) MML-SurgAdapt, a unified task-agnostic CLIP-based model for multi-task surgical analysis; (2) an SPML adaptation that cuts labeling needs by 23\% while enhancing robustness and addressing false negatives; (3) an evaluation of loss functions and benchmarks against task-specific models; and (4) a scalable framework for surgical and broader multimodal applications.

\begin{figure}[t]
    \centering
   \includegraphics[width=0.9\textwidth]{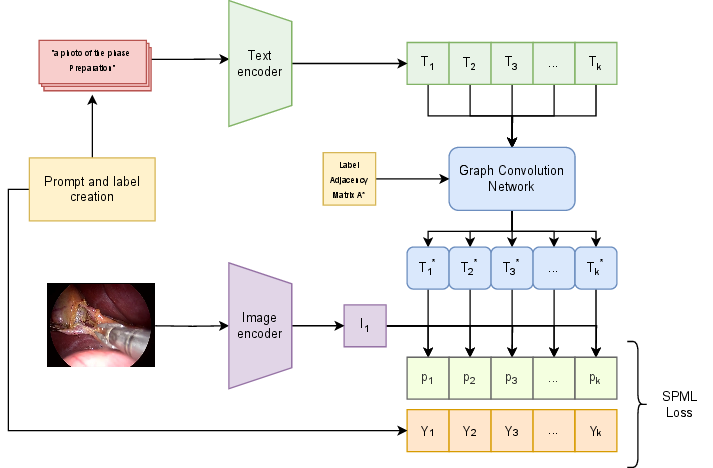}
    \caption{Proposed model architecture of MML-SurgAdapt}
    \label{fig:arch}
\end{figure}
\section{Methodology}
To address multi-task learning in the surgical domain, we use the popular Vision-Language Model (VLM), CLIP \cite{radford2021learning}. For our multi-label classification setup, we convert task-specific labels into textual prompts and calculate the cosine similarity between the image embedding and each label's corresponding text embedding. These similarities are then converted into probabilities, using a sigmoid function to obtain label predictions. 

We validate our approach on the laparoscopic cholecystectomy procedure, focusing on three interrelated tasks: Phase Recognition, Critical View of Safety (CVS) Assessment, and Action Triplet Recognition. 
\subsection{Prompt and label creation}
We create a Partial Positive setup by combining the three datasets, each providing partial annotations for specific tasks (phase, CVS, or triplet). In this setup, each image contains annotations for its respective task, while the labels for other tasks are unobserved and set as negative by default, resulting in false negatives. The ground truth thus contains noise due to these false negatives. The tasks are combined by creating a union of the phase (7 labels), CVS (3 labels), and triplet (100 labels), resulting in 110 labels. This is treated as a multi-label classification problem with missing or unknown data. 

For creation of the SPML setup, we randomly select one positive label per image and set the rest to zero. This setup encourages the model to generalize despite incomplete supervision, learning meaningful representations even with minimal annotations per image. For prompt creation, we include the task and label information in the form of a simple prompt. A general prompt would look like: "a photo of [text for \textit{TASK} and \textit{LABEL}]". Examples include:
\textit{Phase:} "a photo of the phase Calot Triangle Dissection"
\textit{CVS:} "a photo of the lower part of the gallbladder divided from the liver bed to expose the cystic plate"
\textit{Triplet}: "a photo of the tool grasper performing the action grasp on the target gallbladder".


\subsection{Model architecture}

Our model, MML-SurgAdapt, makes use of the CLIP \cite{radford2021learning} framework: the image and text encoders with additional components to adapt to the multi-task SPML classification setup. 
The CLIP image encoder \(f_i(\cdot)\) and text encoder \(f_t(\cdot)\), initialized with pre-trained weights, extract feature representations from the input image and text prompts, respectively. For a given image \(X\) and a pre-defined set of \(K\) text prompts \(\{p_k\}_{k=1}^K\), the encoders generate the following representations:
\begin{equation}
\mathbf{I}_f = f_i(X), \quad \mathbf{T}_f = \{f_t(p_k)\}_{k=1}^K,
\end{equation}
where \(\mathbf{I}_f \in \mathbb{R}^d\) is the image feature vector, and \(\mathbf{T}_f \in \mathbb{R}^{K \times d}\) is the set of text feature vectors for the \(K\) labels, where \(K=110\). Although the text prompts remains fixed, we use only one label per image for backpropagation during training, aligning with the specific requirements of the SPML setup.

In addition, to capture inter-label relationships, we apply a Graph Convolution Network (GCN) \cite{kipf2016semi} on the text features \(\mathbf{T}_f\). This GCN module is used to refine the text features by utilizing the label dependency graph. We represent it in form of an adjacency matrix \(\mathbf{A} = (a_{ij})_{K \times K}\), where \(K\) is the total number of labels. Following \cite{wang2023hierarchical}, we create this matrix using similarities of the text encoder embeddings of the label prompts, and sparsify it to remove noise by retaining only the top-K connections. Finally, to address potential over-smoothing, we introduce a scaling factor to adjust the weights assigned to neighboring label nodes. Let \(\mathbf{H}^{(0)} = \mathbf{T}_f\) be the input to the GCN. The GCN propagates label information as follows:
\begin{equation}
\mathbf{H}^{(l+1)} = \rho\left(\mathbf{A} \mathbf{H}^{(l)} \mathbf{W}^{(l)}\right),
\end{equation}
where \(\mathbf{W}^{(l)}\) are the learnable weights for the \(l\)-th layer, and \(\rho(\cdot)\) is a non-linear activation function. After \(L\) layers, the final text embeddings \(\mathbf{T}_g = \mathbf{H}^{(L)}\) are computed. A residual connection is added, combining the GCN output with the original text embeddings:
\begin{equation}
\mathbf{T}_\text{final} = \mathbf{T}_g + \mathbf{T}_f.
\end{equation}

For multi-label classification, the similarity between the image and each text embedding is calculated using cosine similarity:
\begin{equation}
s_k =\frac{\mathbf{I}_f \cdot \mathbf{T}_\text{final}[k]}{\|\mathbf{I}_f\| \|\mathbf{T}_\text{final}[k]\|},
\end{equation}
where \(s_k\) is the similarity score for the \(k\)-th label. The probabilities for each label are computed as:
\(
p_k = \sigma\left(\frac{s_k}{\tau}\right),
\)
where \(\sigma(z) = \frac{1}{1 + e^{-z}}\) is the sigmoid activation function, and \(\tau\) is a temperature parameter.
Finally, an appropriate loss function \(\mathcal{L}\) is applied to optimize the model. The loss is computed using the ground truth labels \(\{y_{ik}\}\) and the predicted probabilities \(\{p_{ik}\}\), where i corresponds to an instance in the dataset.
The architecture of the proposed model is shown in Fig. \ref{fig:arch}.

We evaluate our framework using different SPML loss functions capable of handling single positive ground truth: WAN \cite{cole2021multi}, SPLC \cite{zhang2021simple} and Hill\cite{zhang2021simple}. Hill Loss outperformed the other loss in all tasks and hence we adapted for our final framework. Hill Loss is designed to be insensitive to missing labels by reweighting the loss for negative labels, assigning less weight to predictions closer to 1, which likely indicate false negatives. For positive labels, a semi-hard mining strategy is utilized, while a modified Mean Squared Error (MSE) loss is applied for negatives, offering better robustness than binary cross-entropy by naturally downweighting high-prediction samples.
The loss functions for positive and negative labels for Hill loss are as follows. 

\begin{align}
p_{km} &= \sigma\left(\frac{s_k}{\tau} - m\right), \quad \text{where } m \text{ is the logit margin}. \\
\text{loss}^+_{\text{Hill}} &= - (1 - p_{km})^\gamma \cdot \log(p_{km}) \\
\text{loss}^-_{\text{Hill}} &= - (\lambda - p_k) \cdot p_k^2, \quad \text{where }  \lambda \text{ is a hyperparameter for reweighting}.
\end{align}

\section{Experimental Setup}

\subsubsection{Dataset splits:} \label{splits}
For each task; Phase recognition, CVS assessment and Action Triplet Recognition, we use publicly available datasets: 
Cholec80 \cite{twinanda2016endonet}, Endoscapes2023 \cite{murali2023endoscapes} and CholecT50 \cite{nwoye2022rendezvous}, respectively.
We combine the datasets for multi-task evaluation using the SPML setup. A detailed description of the datasets, including the number of images for training, validation, and testing, is provided in Tab. \ref{tab:datasets_overview}.  

To ensure a fair evaluation, we enforce mutual exclusivity between the test splits across the datasets. Specifically, there is an overlap between the training and testing images in datasets like Cholec80 and CholecT50. To prevent any data leakage, overlapping videos have been removed from the Cholec80 dataset. As a result, the final split for Cholec80 consists of 36 videos for training, 7 for validation, and 15 for testing. For the Endoscapes2023 dataset, the split is performed according to \cite{murali2023endoscapes}. The CholecT50 dataset uses the Rendezvous split \cite{nwoye2022rendezvous}. 

\begin{table}[ht]
\caption{Overview of datasets, tasks, and evaluation criteria. (T:V:t) represents (Train: Val: Test), I,V,T for triplet recognition represent instrument, verb, target, respectively.}
\label{tab:datasets_overview}
\centering
\resizebox{\textwidth}{!}{%
\begin{tabular}{l | l | c | c | l }
\toprule
\textbf{Dataset} & \textbf{Task} & \textbf{Videos (T:V:t)} & \textbf{Images (T:V:t) }& \textbf{Metrics} \\ 
\midrule
Cholec80  \cite{twinanda2016endonet}       & Phase recognition  & 36:7:15 & 78614:17732:42098  & F1-score                     \\ 
Endoscapes2023 \cite{murali2023endoscapes}      & CVS assessment     & 120:41:40 & 6960:2331:1799   & AP (C1, C2, C3), mAP \\ 
CholecT50  \cite{nwoye2022rendezvous}     & Triplet recognition & 35:5:10 & 72815:6797:21251  & mAPs (I, V, T, IV, IT, IVT) \\ 
\bottomrule
\end{tabular}
}
\end{table}

\noindent \textbf{Evaluation Criteria:} 
During testing, we evaluate the images using only the labels that correspond to the dataset/task from which the images originate. Each task is evaluated separately, ensuring that only the appropriate labels are considered for its assessment. 
For Cholec80-Phase recognition, F1-score is computed according to \cite{alapatt2024jumpstarting}. For Endoscapes2023-CVS assessment, average precision per class (C1, C2, C3), along with the mean average precision (mAP) across all classes is evaluated at frame-level. For CholecT50: Action Triplet recognition, average precision is computed across six configurations (I, V, T, IV, IT, IVT), as described in \cite{nwoye2022rendezvous}.


\noindent \textbf{Implementation details:}
We use the ViT-L/14 based pretrained CLIP model for all our experiments. We train our models for 15 epochs on a batch size of 128. Optimization is done by Adam with a learning rate of 1e-5. All the experiments have been conducted on a single Nvidia A100 GPU. Input images are resized to 224x224 and normalized. The best model is saved based on three validation criteria: highest mAP, lowest loss on partial positive validation ground truths and lowest loss on single positive validation ground truths. The single positive labels are fixed during dataset initialization. We set the hyperparameter for reweighting, \(\lambda \), in the Hill loss equal to 1.5. 

\section{Results and Discussions} \label{results}

The experimental results of the MML-SurgAdapt model are presented across various setups and compared with multiple baselines. 

\subsection{Comparision between Architectures}
\textbf{Vision-only Models:}
For the Endoscapes2023 dataset, we used ResNet50-MoCov2 \cite{murali2023endoscapes} as the benchmark for CVS assessment. This model considers only CVS labels and does not use bounding boxes or segmentation masks. For the CholecT50 dataset, we referenced the Rendezvous model \cite{nwoye2022rendezvous}, as it uses aligned data splits. However, because we trained on a subset of the Cholec80 dataset to avoid overlap, direct comparisons with state-of-the-art (SOTA) methods were not possible.

\noindent \textbf{Task-specific Models:}
We compared our multi-task model against task-specific models that were trained separately on the full labels from each dataset using cross-entropy loss. These baselines include ResNet50, CLIP, and a variant of a multi-task model that uses ResNet50 as a feature extractor with three separate classification heads. Despite training with fewer annotations, our model performed comparable to these task-specific models and established benchmarks.

\noindent \textbf{SPML Models:}
We evaluated the MML-SurgAdapt model in the SPML setup, where each image has only one positive label (fixed during initialization) and all others are negative. We benchmarked MML-SurgAdapt against state-of-the-art SPML frameworks, including VLPL\cite{xing2024vision}, DualCoOp \cite{sun2022dualcoop}, and HSPNet \cite{wang2023hierarchical}. We use the default loss functions as specified in the original papers for each model. MML-SurgAdapt outperforms the existing SPML frameworks for the multi-task integration showing the robustness of the model.

A summary of the comparison against all baselines is provided in Table \ref{tab:combined}. Overall, MML-SurgAdapt achieves competitive performance while significantly reducing labels compared to task-specific models and vision-only baselines.
\begin{table*}[t]
\caption{Comparison against baselines for Phase and CVS Assessment and Action Triplet Recognition. TS stands for Task-specific.}
\label{tab:combined}
\centering
\resizebox{0.95\textwidth}{!}{%
\begin{tabular}{l|c|cccc|cccccc}
\toprule
\textbf{Model} & \textbf{Phase} &  \multicolumn{4}{c|}{\textbf{CVS Assessment}} & \multicolumn{6}{c}{\textbf{Action Triplet Recognition}} \\  
\midrule
 & \textbf{F1} & \textbf{C1} & \textbf{C2} & \textbf{C3} & \textbf{mAP} & \textbf{AP-I} & \textbf{AP-V} & \textbf{AP-T} & \textbf{AP-IV} & \textbf{AP-IT} & \textbf{AP-IVT} \\  
\midrule
\multicolumn{12}{l}{\textbf{Vision-only Models}} \\
\hline
ResNet50-MoCov2 \cite{murali2023endoscapes} & - & 46.4 & \textbf{56.5} & \textbf{69.4} & \textbf{57.4} & - & - & - & - & - & - \\
Rendezvous \cite{nwoye2022rendezvous} & - & - & - & - & - & \textbf{92.0} & 60.7 & 38.3 & \textbf{39.4} & 36.9 & 29.9 \\
\midrule
\multicolumn{12}{l}{\textbf{Task-Specific Models}} \\
\midrule
ResNet50 - TS \cite{he2016deep} & 65.1 & 43.2 & 36.9 & 49.8 & 43.3 & 64.0 & 43.0 & 32.7 & 24.4 & 21.9 & 16.7 \\
CLIPViT-L/14-TS \cite{radford2021learning} & 52.6 & 51.4 & 41.2 & 66.0 & 52.9 & 86.5 & \textbf{62.4} & 41.5 & 39.3 & 35.2 & 29.3 \\
Multi-task ResNet50 & 65.9 & 41.7 & 31.5 & 47.7 & 40.3 & 62.0 & 44.8 & 31.2 & 24.6 & 19.6 & 16.2 \\
\midrule
\multicolumn{12}{l}{\textbf{SPML Models}} \\
\midrule
DualCoOp \cite{sun2022dualcoop} & 40.7 & 37.9& 28.9 & 42.1 & 36.3 & 43.6 & 29.4 & 21.6 & 15.4 & 12.2 & 9.5 \\
VLPL \cite{xing2024vision} & 69.5 & 46.9 & 39.4 & 57.7 & 48.0 & 69.7 & 48.8 & 33.5 & 26.2 & 22.9 & 17.6 \\
HSPNet \cite{wang2023hierarchical} & 68.4 & 45.4 & 41.4 & 55.2 & 47.3 & 77.6 & 55.3 & 37.9 & 32.9 & 29.9 & 23.5 \\
\textbf{MML-SurgAdapt} & \underline{\textbf{74.2}} & \underline{\textbf{55.3}} & \underline{47.0} & \underline{65.4} & \underline{55.9} & \underline{87.3} & \underline{61.1} & \underline{\textbf{43.2}} & \underline{37.8} & \underline{\textbf{37.3}} & \underline{\textbf{30.3}} \\
\bottomrule
\end{tabular}
}
\end{table*}
\subsection{Loss Function Analysis}
We conducted studies across different loss functions and training setups.
We tested WAN \cite{cole2021multi}, SPLC \cite{zhang2021simple}, and Hill Loss \cite{zhang2021simple} in the SPML setup, conducting five runs per experiment to report standard deviations. We then extended the evaluation to the Partial Positive setup, where labels are partially annotated as given in the dataset. Here, the ground truth contains false negatives, but all known labels are marked as positive. Results for the Single and Partial Positive setups in Phase Recognition, CVS Assessment and Action Triplet Recognition are shown in Table \ref{tab:loss_function}.

Notably, WAN slightly outperformed the others in phase and triplet recognition but underperformed in the more complex CVS assessment task, indicating its effectiveness for simpler tasks but limitations in handling complexity. Performance inconsistencies were observed with SPLC and it could stem from its reliance on pseudolabels generated from initial predictions, which can be inaccurate due to the surgical domain's divergence from CLIP's pretraining data. Findings indicate that all three loss functions performed comparably well across tasks, with Hill Loss emerging as the best performer, particularly in CVS assessment. 

In the Partial Positive setup, where additional positive labels were included, we observed only a slight performance gain $\sim$0.5\%, suggesting that the model maintains strong performance even with extreme positive annotations. Quantitatively, the Single Positive setup reduced labels by 23\% compared to the Partial Positive scenario while maintaining competitive performance.
\begin{table*}[t]
\caption{Combined Performance comparison of objective functions in SPML and Partial Positive setups: Phase, CVS Assessment and Action Triplet Recognition}
\label{tab:loss_function}
\centering
\resizebox{0.9\textwidth}{!}{%
\begin{tabular}{l|c|c|cccccc}
\toprule
\textbf{Loss} & \textbf{Phase} &  \multicolumn{1}{c|}{\textbf{CVS}} & \multicolumn{6}{c}{\textbf{Action Triplet Recognition}} \\  
\midrule
 & \textbf{F1}  & \textbf{mAP} & \textbf{AP-I} & \textbf{AP-V} & \textbf{AP-T} & \textbf{AP-IV} & \textbf{AP-IT} & \textbf{AP-IVT} \\  
\midrule
\multicolumn{9}{l}{\textbf{SPML Setup}} \\
\midrule
WAN \cite{cole2021multi} & \textbf{74.7±0.7} & 47.8±1.1  & \textbf{90.2±0.3} & \textbf{63.1±0.4} & \textbf{45.3±0.8} & \textbf{40.9±0.4} & \textbf{37.3±0.9} & \textbf{30.9±0.5} \\
SPLC \cite{zhang2021simple} & 73.2±0.6 & 54.0±2.4 & 87.5±0.9 & 62.1±1.1 & 42.5±1.0 & 38.4±1.2 & 36.3±1.0   & 29.5±0.9  \\
Hill \cite{zhang2021simple} & 74.2±0.6 & \textbf{55.9±2.8} & 87.3±1.2 & 61.1±1.4 & 43.2±1.3 & 37.8±1.4 & 37.3±0.5   & 30.3±0.4 \\
\midrule
\multicolumn{9}{l}{\textbf{Partial Positive Setup}} \\
\midrule
WAN \cite{cole2021multi} & \textbf{75.1±0.9} & 51.3±1.4 &  \textbf{89.9±0.5}  & 62.9±0.6 & \textbf{46.2±1.2} & 41.6±0.9 & 36.6±0.8 & 31.1±0.6 \\
SPLC \cite{zhang2021simple} & 73.4±0.3 & 53.1±2.8 & 88.4±0.9 & 63.0±0.8 & 43.7±1.2 & 40.1±0.6 & 36.7±1.9 & 30.1±1.1 \\
Hill \cite{zhang2021simple} & 74.2±0.8 & \textbf{54.9±3.0} & 89.4±1.0 & \textbf{64.6±1.1} & 45.4±0.2 & \textbf{41.6±1.1} & \textbf{38.0±1.5}   & \textbf{31.6±1.3}  \\
\bottomrule
\end{tabular}
}
\end{table*}

\subsection{Limitations and Future Work}

An ablation study removing the Graph Convolutional Network (GCN) showed only a 0.5\% performance improvement when included. While the full model requires end-to-end fine-tuning due to its use of pretrained weights, this limits the performance improvement of the GCN. Differences may arise with changes in initialization; in this work, we used CLIP weights, but a domain-specific pretrained model, particularly one trained on surgical data, could enable more effective adaptation of the GCN without full fine-tuning. While only basic initial tests were conducted, exploring this remains future work.

The current setup also does not explicitly address class imbalance. Preliminary experiments with class weighting yielded inconclusive results and were omitted, but targeted strategies to handle imbalance, especially for rare action-triplets, remain to be explored. Lastly, our evaluation is limited to a single procedure, and validating the framework across a broader range of surgeries is an important direction for future work.

\section{Conclusion}

In this work, we addressed the limitations of traditional approaches in Surgical AI, which rely on task specific datasets and models that lack scalability and struggle to integrate diverse tasks. We introduce MML-SurgAdapt, a unified, task-agnostic framework that uses natural language supervision to align surgical images with text prompts across multiple tasks. By incorporating Single Positive Multi-Label (SPML) learning in the surgical domain, we tackle the challenges of combining datasets for interrelated tasks, mitigating false negatives, and enabling label-efficient learning from partial annotations. Through extensive experiments, we compare our model against existing task-specific baselines, and SPML frameworks, while also analyzing the impact of different loss functions in handling false negatives. Our findings demonstrate that a single AI model can generalize across diverse complementary surgical tasks within a procedure, paving the way for future advancements in surgical AI and annotation-efficient learning.

\begin{credits}
\subsubsection{\ackname}

This work has received funding from the European Union (ERC, CompSURG, 101088553). Views and opinions expressed are however those of the authors only and do not necessarily reflect those of the European Union or the European Research Council. Neither the European Union nor the granting authority can be held responsible for them. This work was also partially supported by French state funds managed by the ANR under Grant ANR-10-IAHU-02. The authors would like to acknowledge the High Performance Computing Center of the University of Strasbourg for supporting this work by providing scientific support and access to computing resources. Part of the computing resources were funded by the Equipex Equip@Meso project (Programme Investissements d’Avenir) and the CPER Alsacalcul/Big Data.

\subsubsection{\discintname}
The authors have no competing interests to declare relevant to this article.

\end{credits}
%
%
%
%

\end{document}